\documentclass[sn-mathphys,Numbered,]{sn-jnl}

\usepackage{graphicx}%
\usepackage{multirow}%
\usepackage{amsmath,amssymb,amsfonts}%
\usepackage{amsthm}%
\usepackage{mathrsfs}%
\usepackage[title]{appendix}%
\usepackage{xcolor}%
\usepackage{textcomp}%
\usepackage{manyfoot}%
\usepackage{booktabs}%
\usepackage{algorithm}%
\usepackage{algorithmicx}%
\usepackage{algpseudocode}%
\usepackage{listings}%
\usepackage{natbib}

\begin{document}

\title[Article Title]{Fixing the problems of deep neural networks will require better training data and learning algorithms}


\author*[1,2]{\fnm{Drew} \sur{Linsley}}\email{drew\_linsley@brown.edu}

\author[1,2,3]{\fnm{Thomas} \sur{Serre}}\email{thomas\_serre@brown.edu}

\affil[1]{\orgdiv{Department of Cognitive, Linguistic, \& Psychological Sciences}, \orgname{Brown University}, \orgaddress{\city{Providence}, \state{RI}}}

\affil[2]{\orgdiv{Carney Institute for Brain Science}, \orgname{Brown University}, \orgaddress{\city{Providence}, \state{RI}}}

\affil[3]{\orgname{Artificial and Natural Intelligence Toulouse Institute}, \orgaddress{\city{Toulouse}, \country{France}}}


\abstract{Bowers and colleagues argue that DNNs are poor models of biological vision because they often learn to rival human accuracy by relying on strategies that differ markedly from those of humans. We show that this problem is worsening as DNNs are becoming larger-scale and increasingly more accurate, and prescribe methods for building DNNs that can reliably model biological vision.}




\maketitle

\section{Introduction}\label{sec1}
Over the past decade, vision scientists have turned to deep neural networks (DNNs) to model biological vision. The popularity of DNNs comes from their ability to rival human performance on visual tasks~\cite{Geirhos2021-rr} and the seemingly concomitant correspondence of their hidden units with biological vision~\cite{Yamins2014-ba}. Bowers and colleagues~\cite{Bowers2022-cy} marshal evidence from psychology and neuroscience to argue that while DNNs and biological systems may achieve similar accuracy on visual benchmarks, they often do so by relying on qualitatively different visual features and strategies~\cite{Malhotra2022-dl,Malhotra2020-kj,Baker2018-wg}. Based on these findings, Bowers and colleagues call for a re-evaluation of what DNNs can tell us about biological vision and suggest dramatic adjustments going forward, potentially even moving on from DNNs altogether. Are DNNs poorly suited to model biological vision?

\begin{figure}[h!]
\begin{center}
\includegraphics[width=.99\linewidth]{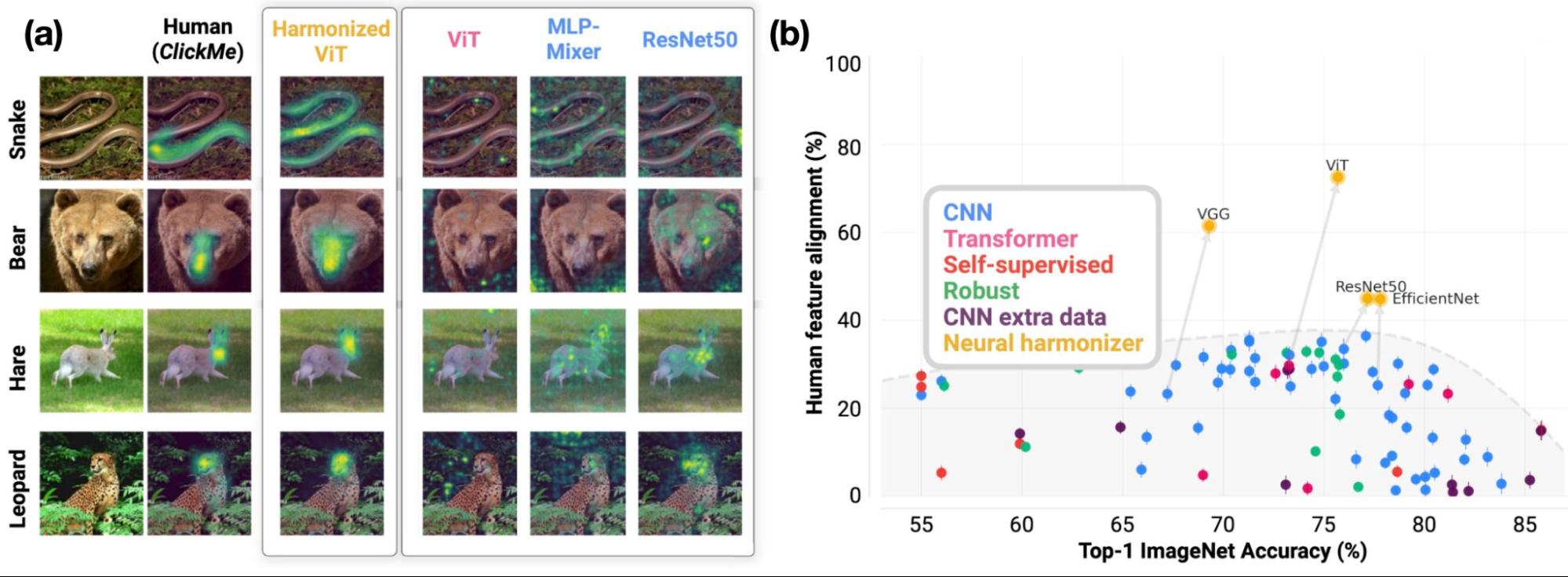}
\end{center}
   \caption{\textbf{Figure 1. A growing misalignment between DNNs and biological vision (adapted from Fel* et al., 2023).} (a) Diagnostic features for object classification differ between humans and DNNs. (b) The spearman correlation between human and DNN feature importance maps has begun to decrease as a function of DNN accuracy on ImageNet.}
\label{fig:qualitative}
\end{figure}

Systematically evaluating DNNs for biological vision. While this review identifies multiple shortcuts in DNNs that are commonly used in vision science, such as ResNet and AlexNet, it does not delve into the root causes of these issues or how widespread these are across different DNN architectures and training routines. We addressed these questions with ClickMe, a web-based game in which human participants teach DNNs how to recognize objects by highlighting category-diagnostic visual features~\cite{Linsley2017-qe,Linsley2019-xi}. With ClickMe, we collected annotations of the visual features that humans rely on to recognize approximately 25\% of ImageNet images (\url{https://serre-lab.github.io/Harmonization/}). Human feature importance maps from ClickMe reveal startling regularity: animals were categorized by their faces, whereas inanimate objects like cars were categorized by their wheels and headlights (Fig.~\ref{fig:qualitative}a). Human participants were also significantly more accurate at rapid object classification when basing their decisions on these features rather than image saliency. In contrast, while DNNs sometimes selected the same diagnostic features as humans, they often relied on “shortcuts” for object recognition~\cite{Geirhos2020-nl}. For example, a DNN called the Vision transformer (ViT) relied on background features, like grass, to recognize a hare, whereas human participants focused almost exclusively on the animal’s head (Fig. 1a). Even more concerning is that the visual features and strategies of humans and DNNs are becoming increasingly misaligned as newer DNNs become more accurate (Fig. 1b). We and others have observed similar trade-offs between DNN accuracy on ImageNet and their ability to explain various human behavioral data and psychophysics~\cite{Fel2022-og,Kumar2022-lv}. Our work indicates that the mismatch between DNN and biological vision identified by Bowers and colleagues is pervasive and worsening.

The next generation of DNNs for biological vision. Bowers and colleagues argue that the inability of DNNs to learn human-like visual strategies reflects architectural limitations. They are correct that there is a rich literature demonstrating how mechanisms inspired by Neuroscience can improve the capabilities of DNNs, helping them learn perceptual grouping~\cite{Kim2020-yw,Linsley2018-ls,Linsley2020-en,Kim2018-qr}, visual reasoning~\cite{Vaishnav2022-hy}, robust object recognition~\cite{Dapello2023-uk}, and to more accurately predict neural activity~\cite{Kubilius2018-gd,Nayebi2018-dc,Bakhtiari2021-yt}. The other fundamental difference between DNNs and biological organisms is how they learn; humans and DNNs learn from vastly different types of data with presumably different objective functions. We believe that the limitations raised by Bowers and colleagues result from a mismatch in data diets and objective functions because we significantly improved the alignment of state-of-the-art DNNs with humans by introducing ClickMe data into their training routines (“Neural harmonizer,” Fig.~\ref{fig:qualitative}). 

Biologically-inspired data diets and objective functions. We believe that the power of DNNs for biological vision is from their ability to generate computational- and algorithmic-level hypotheses about vision, which will guide experiments to identify plausible circuits. For instance, the great success of gradient descent and backpropagation for training DNNs has inspired the search for biologically-plausible approximations~\cite{Lillicrap2020-hg}. Visual Neuroscience is similarly positioned to benefit from DNNs if we can improve their alignment with biology.

The most straightforward opportunity for aligning DNNs with biological vision is to train them with more biologically-plausible data and objective functions~\cite{Smith2017-rz}. There have been efforts to do this with first-person video, however, these efforts have failed to yield much benefit in computer vision or other aspects of biological vision~\cite{Orhan2020-pk,Sullivan2021-wy,Zhuang2021-ds}, potentially because the small scale of these datasets makes them ill-suited for training DNNs. An alternative approach is to utilize advances in 3D computer vision, like Neural Radiance Fields~\cite{Mildenhall2020-sy}, to generate spatiotemporal (stereo) datasets for training DNNs that are infinitely scalable and can be integrated with other modalities, such as somatosensation and language. It is also very likely that objective functions that will lead to human-like visual strategies and features from these datasets have yet to be discovered. However, promising directions include optimizing for slow feature analysis~\cite{Wiskott2002-au} and predictive coding~\cite{Lotter2016-dw,Mineault2021-xc}, which could help align DNNs with humans without relying on ClickMe data.

Aligned DNNs may be all we need. Bowers and colleagues point out a number of ways in which DNNs fail as models of biological vision. These problems are pervasive and likely caused by the standard image datasets and training routines of DNNs, which are guided by Engineering rather than Biology. Bowers and colleagues may well be right that an entirely new class of models is needed to account for biological vision, but at the moment there are no viable alternatives. Until other model classes can rival human performance on visual tasks, we suspect that the most productive path forward towards modeling biological vision and aligning DNNs with biological vision is to develop more biologically-plausible data diets and objective functions.

\paragraph{Acknowledgements.} We thank Lakshmi Govindarajan for his helpful comments and feedback on drafts of this commentary.

\paragraph{Competing interests.} The authors of this commentary have no competing interests.

\paragraph{Funding.} This work was supported by ONR (N00014-19-1-2029), NSF (IIS-1912280) and the ANR-3IA Artificial and Natural Intelligence Toulouse Institute (ANR-19-PI3A-0004).

\bibliography{sn-bibliography}


\begin{thebibliography}{29}
\ifx \bisbn   \undefined \def \bisbn  #1{ISBN #1}\fi
\ifx \binits  \undefined \def \binits#1{#1}\fi
\ifx \bauthor  \undefined \def \bauthor#1{#1}\fi
\ifx \batitle  \undefined \def \batitle#1{#1}\fi
\ifx \bjtitle  \undefined \def \bjtitle#1{#1}\fi
\ifx \bvolume  \undefined \def \bvolume#1{\textbf{#1}}\fi
\ifx \byear  \undefined \def \byear#1{#1}\fi
\ifx \bissue  \undefined \def \bissue#1{#1}\fi
\ifx \bfpage  \undefined \def \bfpage#1{#1}\fi
\ifx \blpage  \undefined \def \blpage #1{#1}\fi
\ifx \burl  \undefined \def \burl#1{\textsf{#1}}\fi
\ifx \doiurl  \undefined \def \doiurl#1{\url{https://doi.org/#1}}\fi
\ifx \betal  \undefined \def \betal{\textit{et al.}}\fi
\ifx \binstitute  \undefined \def \binstitute#1{#1}\fi
\ifx \binstitutionaled  \undefined \def \binstitutionaled#1{#1}\fi
\ifx \bctitle  \undefined \def \bctitle#1{#1}\fi
\ifx \beditor  \undefined \def \beditor#1{#1}\fi
\ifx \bpublisher  \undefined \def \bpublisher#1{#1}\fi
\ifx \bbtitle  \undefined \def \bbtitle#1{#1}\fi
\ifx \bedition  \undefined \def \bedition#1{#1}\fi
\ifx \bseriesno  \undefined \def \bseriesno#1{#1}\fi
\ifx \blocation  \undefined \def \blocation#1{#1}\fi
\ifx \bsertitle  \undefined \def \bsertitle#1{#1}\fi
\ifx \bsnm \undefined \def \bsnm#1{#1}\fi
\ifx \bsuffix \undefined \def \bsuffix#1{#1}\fi
\ifx \bparticle \undefined \def \bparticle#1{#1}\fi
\ifx \barticle \undefined \def \barticle#1{#1}\fi
\bibcommenthead
\ifx \bconfdate \undefined \def \bconfdate #1{#1}\fi
\ifx \botherref \undefined \def \botherref #1{#1}\fi
\ifx \url \undefined \def \url#1{\textsf{#1}}\fi
\ifx \bchapter \undefined \def \bchapter#1{#1}\fi
\ifx \bbook \undefined \def \bbook#1{#1}\fi
\ifx \bcomment \undefined \def \bcomment#1{#1}\fi
\ifx \oauthor \undefined \def \oauthor#1{#1}\fi
\ifx \citeauthoryear \undefined \def \citeauthoryear#1{#1}\fi
\ifx \endbibitem  \undefined \def \endbibitem {}\fi
\ifx \bconflocation  \undefined \def \bconflocation#1{#1}\fi
\ifx \arxivurl  \undefined \def \arxivurl#1{\textsf{#1}}\fi
\csname PreBibitemsHook\endcsname

\bibitem[\protect\citeauthoryear{Geirhos et~al.}{2021}]{Geirhos2021-rr}
\begin{botherref}
\oauthor{\bsnm{Geirhos}, \binits{R.}},
\oauthor{\bsnm{Narayanappa}, \binits{K.}},
\oauthor{\bsnm{Mitzkus}, \binits{B.}},
\oauthor{\bsnm{Thieringer}, \binits{T.}},
\oauthor{\bsnm{Bethge}, \binits{M.}},
\oauthor{\bsnm{Wichmann}, \binits{F.A.}},
\oauthor{\bsnm{Brendel}, \binits{W.}}:
Partial success in closing the gap between human and machine vision
(2021)
{\href{https://arxiv.org/abs/2106.07411}{{arXiv:2106.07411}}}
{[cs.CV]}
\end{botherref}
\endbibitem

\bibitem[\protect\citeauthoryear{Yamins et~al.}{2014}]{Yamins2014-ba}
\begin{barticle}
\bauthor{\bsnm{Yamins}, \binits{D.L.K.}},
\bauthor{\bsnm{Hong}, \binits{H.}},
\bauthor{\bsnm{Cadieu}, \binits{C.F.}},
\bauthor{\bsnm{Solomon}, \binits{E.A.}},
\bauthor{\bsnm{Seibert}, \binits{D.}},
\bauthor{\bsnm{DiCarlo}, \binits{J.J.}}:
\batitle{Performance-optimized hierarchical models predict neural responses in higher visual cortex}.
\bjtitle{Proc. Natl. Acad. Sci. U. S. A.}
\bvolume{111}(\bissue{23}),
\bfpage{8619}--\blpage{8624}
(\byear{2014})
\end{barticle}
\endbibitem

\bibitem[\protect\citeauthoryear{Bowers et~al.}{2022}]{Bowers2022-cy}
\begin{botherref}
\oauthor{\bsnm{Bowers}, \binits{J.S.}},
\oauthor{\bsnm{Malhotra}, \binits{G.}},
\oauthor{\bsnm{Dujmovi{\'c}}, \binits{M.}},
\oauthor{\bsnm{Montero}, \binits{M.L.}},
\oauthor{\bsnm{Tsvetkov}, \binits{C.}},
\oauthor{\bsnm{Biscione}, \binits{V.}},
\oauthor{\bsnm{Puebla}, \binits{G.}},
\oauthor{\bsnm{Adolfi}, \binits{F.}},
\oauthor{\bsnm{Hummel}, \binits{J.E.}},
\oauthor{\bsnm{Heaton}, \binits{R.F.}},
\oauthor{\bsnm{Evans}, \binits{B.D.}},
\oauthor{\bsnm{Mitchell}, \binits{J.}},
\oauthor{\bsnm{Blything}, \binits{R.}}:
Deep problems with neural network models of human vision.
Behav. Brain Sci.,
1--74
(2022)
\end{botherref}
\endbibitem

\bibitem[\protect\citeauthoryear{Malhotra et~al.}{2022}]{Malhotra2022-dl}
\begin{barticle}
\bauthor{\bsnm{Malhotra}, \binits{G.}},
\bauthor{\bsnm{Dujmovi{\'c}}, \binits{M.}},
\bauthor{\bsnm{Bowers}, \binits{J.S.}}:
\batitle{Feature blindness: A challenge for understanding and modelling visual object recognition}.
\bjtitle{PLoS Comput. Biol.}
\bvolume{18}(\bissue{5}),
\bfpage{1009572}
(\byear{2022})
\end{barticle}
\endbibitem

\bibitem[\protect\citeauthoryear{Malhotra et~al.}{2020}]{Malhotra2020-kj}
\begin{barticle}
\bauthor{\bsnm{Malhotra}, \binits{G.}},
\bauthor{\bsnm{Evans}, \binits{B.D.}},
\bauthor{\bsnm{Bowers}, \binits{J.S.}}:
\batitle{Hiding a plane with a pixel: examining shape-bias in {CNNs} and the benefit of building in biological constraints}.
\bjtitle{Vision Res.}
\bvolume{174},
\bfpage{57}--\blpage{68}
(\byear{2020})
\end{barticle}
\endbibitem

\bibitem[\protect\citeauthoryear{Baker et~al.}{2018}]{Baker2018-wg}
\begin{barticle}
\bauthor{\bsnm{Baker}, \binits{N.}},
\bauthor{\bsnm{Lu}, \binits{H.}},
\bauthor{\bsnm{Erlikhman}, \binits{G.}},
\bauthor{\bsnm{Kellman}, \binits{P.J.}}:
\batitle{Deep convolutional networks do not classify based on global object shape}.
\bjtitle{PLoS Comput. Biol.}
\bvolume{14}(\bissue{12}),
\bfpage{1006613}
(\byear{2018})
\end{barticle}
\endbibitem

\bibitem[\protect\citeauthoryear{Linsley et~al.}{2017}]{Linsley2017-qe}
\begin{bchapter}
\bauthor{\bsnm{Linsley}, \binits{D.}},
\bauthor{\bsnm{Eberhardt}, \binits{S.}},
\bauthor{\bsnm{Sharma}, \binits{T.}},
\bauthor{\bsnm{Gupta}, \binits{P.}},
\bauthor{\bsnm{Serre}, \binits{T.}}:
\bctitle{What are the visual features underlying human versus machine vision?}
In: \bbtitle{2017 {IEEE} International Conference on Computer Vision Workshops ({ICCVW})},
pp. \bfpage{2706}--\blpage{2714}
(\byear{2017})
\end{bchapter}
\endbibitem

\bibitem[\protect\citeauthoryear{Linsley et~al.}{2019}]{Linsley2019-xi}
\begin{botherref}
\oauthor{\bsnm{Linsley}, \binits{D.}},
\oauthor{\bsnm{Shiebler}, \binits{D.}},
\oauthor{\bsnm{Eberhardt}, \binits{S.}},
\oauthor{\bsnm{Serre}, \binits{T.}}:
Learning what and where to attend.
International Conference on Learning Representations (ICLR)
(2019)
\end{botherref}
\endbibitem

\bibitem[\protect\citeauthoryear{Geirhos et~al.}{2020}]{Geirhos2020-nl}
\begin{barticle}
\bauthor{\bsnm{Geirhos}, \binits{R.}},
\bauthor{\bsnm{Jacobsen}, \binits{J.-H.}},
\bauthor{\bsnm{Michaelis}, \binits{C.}},
\bauthor{\bsnm{Zemel}, \binits{R.}},
\bauthor{\bsnm{Brendel}, \binits{W.}},
\bauthor{\bsnm{Bethge}, \binits{M.}},
\bauthor{\bsnm{Wichmann}, \binits{F.A.}}:
\batitle{Shortcut learning in deep neural networks}.
\bjtitle{Nature Machine Intelligence}
\bvolume{2}(\bissue{11}),
\bfpage{665}--\blpage{673}
(\byear{2020})
\end{barticle}
\endbibitem

\bibitem[\protect\citeauthoryear{Fel* et~al.}{2022}]{Fel2022-og}
\begin{botherref}
\oauthor{\bsnm{Fel*}, \binits{T.}},
\oauthor{\bsnm{Felipe*}, \binits{I.}},
\oauthor{\bsnm{Linsley*}, \binits{D.}},
\oauthor{\bsnm{Serre}, \binits{T.}}:
Harmonizing the object recognition strategies of deep neural networks with humans.
Adv. Neural Inf. Process. Syst.
(2022)
\end{botherref}
\endbibitem

\bibitem[\protect\citeauthoryear{Kumar et~al.}{2022}]{Kumar2022-lv}
\begin{botherref}
\oauthor{\bsnm{Kumar}, \binits{M.}},
\oauthor{\bsnm{Houlsby}, \binits{N.}},
\oauthor{\bsnm{Kalchbrenner}, \binits{N.}},
\oauthor{\bsnm{Cubuk}, \binits{E.D.}}:
Do better {ImageNet} classifiers assess perceptual similarity better?
(2022)
\end{botherref}
\endbibitem

\bibitem[\protect\citeauthoryear{Kim* et~al.}{2020}]{Kim2020-yw}
\begin{botherref}
\oauthor{\bsnm{Kim*}, \binits{J.}},
\oauthor{\bsnm{Linsley*}, \binits{D.}},
\oauthor{\bsnm{Thakkar}, \binits{K.}},
\oauthor{\bsnm{Serre}, \binits{T.}}:
Disentangling neural mechanisms for perceptual grouping.
International Conference on Representation Learning
(2020)
\end{botherref}
\endbibitem

\bibitem[\protect\citeauthoryear{Linsley et~al.}{2018}]{Linsley2018-ls}
\begin{botherref}
\oauthor{\bsnm{Linsley}, \binits{D.}},
\oauthor{\bsnm{Kim}, \binits{J.}},
\oauthor{\bsnm{Veerabadran}, \binits{V.}},
\oauthor{\bsnm{Serre}, \binits{T.}}:
Learning long-range spatial dependencies with horizontal gated-recurrent units
(2018)
{\href{https://arxiv.org/abs/1805.08315}{{arXiv:1805.08315}}}
{[cs.CV]}
\end{botherref}
\endbibitem

\bibitem[\protect\citeauthoryear{Linsley et~al.}{2020}]{Linsley2020-en}
\begin{botherref}
\oauthor{\bsnm{Linsley}, \binits{D.}},
\oauthor{\bsnm{Kim}, \binits{J.}},
\oauthor{\bsnm{Ashok}, \binits{A.}},
\oauthor{\bsnm{Serre}, \binits{T.}}:
Recurrent neural circuits for contour detection.
International Conference on Learning Representations
(2020)
\end{botherref}
\endbibitem

\bibitem[\protect\citeauthoryear{Kim et~al.}{2018}]{Kim2018-qr}
\begin{barticle}
\bauthor{\bsnm{Kim}, \binits{J.}},
\bauthor{\bsnm{Ricci}, \binits{M.}},
\bauthor{\bsnm{Serre}, \binits{T.}}:
\batitle{{Not-So-CLEVR}: learning same-different relations strains feedforward neural networks}.
\bjtitle{Interface Focus}
\bvolume{8}(\bissue{4}),
\bfpage{20180011}
(\byear{2018})
\end{barticle}
\endbibitem

\bibitem[\protect\citeauthoryear{Vaishnav et~al.}{2022}]{Vaishnav2022-hy}
\begin{barticle}
\bauthor{\bsnm{Vaishnav}, \binits{M.}},
\bauthor{\bsnm{Cadene}, \binits{R.}},
\bauthor{\bsnm{Alamia}, \binits{A.}},
\bauthor{\bsnm{Linsley}, \binits{D.}},
\bauthor{\bsnm{VanRullen}, \binits{R.}},
\bauthor{\bsnm{Serre}, \binits{T.}}:
\batitle{Understanding the computational demands underlying visual reasoning}.
\bjtitle{Neural Comput.}
\bvolume{34}(\bissue{5}),
\bfpage{1075}--\blpage{1099}
(\byear{2022})
\end{barticle}
\endbibitem

\bibitem[\protect\citeauthoryear{Dapello et~al.}{2023}]{Dapello2023-uk}
\begin{botherref}
\oauthor{\bsnm{Dapello}, \binits{J.}},
\oauthor{\bsnm{Kar}, \binits{K.}},
\oauthor{\bsnm{Schrimpf}, \binits{M.}},
\oauthor{\bsnm{Geary}, \binits{R.B.}},
\oauthor{\bsnm{Ferguson}, \binits{M.}},
\oauthor{\bsnm{Cox}, \binits{D.D.}},
\oauthor{\bsnm{DiCarlo}, \binits{J.J.}}:
Aligning Model and Macaque Inferior Temporal Cortex Representations Improves {Model-to-Human} Behavioral Alignment and Adversarial Robustness
(2023)
\end{botherref}
\endbibitem

\bibitem[\protect\citeauthoryear{Kubilius et~al.}{2018}]{Kubilius2018-gd}
\begin{botherref}
\oauthor{\bsnm{Kubilius}, \binits{J.}},
\oauthor{\bsnm{Schrimpf}, \binits{M.}},
\oauthor{\bsnm{Nayebi}, \binits{A.}},
\oauthor{\bsnm{Bear}, \binits{D.}},
\oauthor{\bsnm{Yamins}, \binits{D.L.K.}},
\oauthor{\bsnm{DiCarlo}, \binits{J.J.}}:
{CORnet}: Modeling the Neural Mechanisms of Core Object Recognition
(2018)
\end{botherref}
\endbibitem

\bibitem[\protect\citeauthoryear{Nayebi et~al.}{2018}]{Nayebi2018-dc}
\begin{botherref}
\oauthor{\bsnm{Nayebi}, \binits{A.}},
\oauthor{\bsnm{Bear}, \binits{D.}},
\oauthor{\bsnm{Kubilius}, \binits{J.}},
\oauthor{\bsnm{Kar}, \binits{K.}},
\oauthor{\bsnm{Ganguli}, \binits{S.}},
\oauthor{\bsnm{Sussillo}, \binits{D.}},
\oauthor{\bsnm{DiCarlo}, \binits{J.J.}},
\oauthor{\bsnm{Yamins}, \binits{D.L.K.}}:
{Task-Driven} convolutional recurrent models of the visual system
(2018)
{\href{https://arxiv.org/abs/1807.00053}{{arXiv:1807.00053}}}
{[q-bio.NC]}
\end{botherref}
\endbibitem

\bibitem[\protect\citeauthoryear{Bakhtiari et~al.}{2021}]{Bakhtiari2021-yt}
\begin{barticle}
\bauthor{\bsnm{Bakhtiari}, \binits{S.}},
\bauthor{\bsnm{Mineault}, \binits{P.}},
\bauthor{\bsnm{Lillicrap}, \binits{T.}},
\bauthor{\bsnm{Pack}, \binits{C.}},
\bauthor{\bsnm{Richards}, \binits{B.}}:
\batitle{The functional specialization of visual cortex emerges from training parallel pathways with self-supervised predictive learning}.
\bjtitle{Adv. Neural Inf. Process. Syst.}
\bvolume{34},
\bfpage{25164}--\blpage{25178}
(\byear{2021})
\end{barticle}
\endbibitem

\bibitem[\protect\citeauthoryear{Lillicrap et~al.}{2020}]{Lillicrap2020-hg}
\begin{barticle}
\bauthor{\bsnm{Lillicrap}, \binits{T.P.}},
\bauthor{\bsnm{Santoro}, \binits{A.}},
\bauthor{\bsnm{Marris}, \binits{L.}},
\bauthor{\bsnm{Akerman}, \binits{C.J.}},
\bauthor{\bsnm{Hinton}, \binits{G.}}:
\batitle{Backpropagation and the brain}.
\bjtitle{Nat. Rev. Neurosci.}
\bvolume{21}(\bissue{6}),
\bfpage{335}--\blpage{346}
(\byear{2020})
\end{barticle}
\endbibitem

\bibitem[\protect\citeauthoryear{Smith and Slone}{2017}]{Smith2017-rz}
\begin{barticle}
\bauthor{\bsnm{Smith}, \binits{L.B.}},
\bauthor{\bsnm{Slone}, \binits{L.K.}}:
\batitle{A developmental approach to machine learning?}
\bjtitle{Front. Psychol.}
\bvolume{8},
\bfpage{2124}
(\byear{2017})
\end{barticle}
\endbibitem

\bibitem[\protect\citeauthoryear{Orhan et~al.}{}]{Orhan2020-pk}
\begin{botherref}
\oauthor{\bsnm{Orhan}, \binits{E.}},
\oauthor{\bsnm{Gupta}, \binits{V.}},
\oauthor{\bsnm{Lake}, \binits{B.M.}}:
Self-supervised learning through the eyes of a child.
In: Larochelle, H., Ranzato, M., Hadsell, R., Balcan, M.F., Lin, H. (eds.)
Advances in Neural Information Processing Systems.
Curran Associates, Inc.
\end{botherref}
\endbibitem

\bibitem[\protect\citeauthoryear{Sullivan et~al.}{2021}]{Sullivan2021-wy}
\begin{barticle}
\bauthor{\bsnm{Sullivan}, \binits{J.}},
\bauthor{\bsnm{Mei}, \binits{M.}},
\bauthor{\bsnm{Perfors}, \binits{A.}},
\bauthor{\bsnm{Wojcik}, \binits{E.}},
\bauthor{\bsnm{Frank}, \binits{M.C.}}:
\batitle{{SAYCam}: A large, longitudinal audiovisual dataset recorded from the infant's perspective}.
\bjtitle{Open Mind (Camb)}
\bvolume{5},
\bfpage{20}--\blpage{29}
(\byear{2021})
\end{barticle}
\endbibitem

\bibitem[\protect\citeauthoryear{Zhuang et~al.}{2021}]{Zhuang2021-ds}
\begin{botherref}
\oauthor{\bsnm{Zhuang}, \binits{C.}},
\oauthor{\bsnm{Yan}, \binits{S.}},
\oauthor{\bsnm{Nayebi}, \binits{A.}},
\oauthor{\bsnm{Schrimpf}, \binits{M.}},
\oauthor{\bsnm{Frank}, \binits{M.C.}},
\oauthor{\bsnm{DiCarlo}, \binits{J.J.}},
\oauthor{\bsnm{Yamins}, \binits{D.L.K.}}:
Unsupervised neural network models of the ventral visual stream.
Proc. Natl. Acad. Sci. U. S. A.
\textbf{118}(3)
(2021)
\end{botherref}
\endbibitem

\bibitem[\protect\citeauthoryear{Mildenhall et~al.}{2020}]{Mildenhall2020-sy}
\begin{botherref}
\oauthor{\bsnm{Mildenhall}, \binits{B.}},
\oauthor{\bsnm{Srinivasan}, \binits{P.P.}},
\oauthor{\bsnm{Tancik}, \binits{M.}},
\oauthor{\bsnm{Barron}, \binits{J.T.}},
\oauthor{\bsnm{Ramamoorthi}, \binits{R.}},
\oauthor{\bsnm{Ng}, \binits{R.}}:
{NeRF}: Representing scenes as neural radiance fields for view synthesis
(2020)
{\href{https://arxiv.org/abs/2003.08934}{{arXiv:2003.08934}}}
{[cs.CV]}
\end{botherref}
\endbibitem

\bibitem[\protect\citeauthoryear{Wiskott and Sejnowski}{2002}]{Wiskott2002-au}
\begin{barticle}
\bauthor{\bsnm{Wiskott}, \binits{L.}},
\bauthor{\bsnm{Sejnowski}, \binits{T.J.}}:
\batitle{Slow feature analysis: unsupervised learning of invariances}.
\bjtitle{Neural Comput.}
\bvolume{14}(\bissue{4}),
\bfpage{715}--\blpage{770}
(\byear{2002})
\end{barticle}
\endbibitem

\bibitem[\protect\citeauthoryear{Lotter et~al.}{2016}]{Lotter2016-dw}
\begin{botherref}
\oauthor{\bsnm{Lotter}, \binits{W.}},
\oauthor{\bsnm{Kreiman}, \binits{G.}},
\oauthor{\bsnm{Cox}, \binits{D.}}:
Deep predictive coding networks for video prediction and unsupervised learning
(2016)
{\href{https://arxiv.org/abs/1605.08104}{{arXiv:1605.08104}}}
{[cs.LG]}
\end{botherref}
\endbibitem

\bibitem[\protect\citeauthoryear{Mineault et~al.}{}]{Mineault2021-xc}
\begin{botherref}
\oauthor{\bsnm{Mineault}, \binits{P.}},
\oauthor{\bsnm{Bakhtiari}, \binits{S.}},
\oauthor{\bsnm{Richards}, \binits{B.}},
\oauthor{\bsnm{Pack}, \binits{C.}}:
Your head is there to move you around: Goal-driven models of the primate dorsal pathway.
In: Ranzato, M., Beygelzimer, A., Dauphin, Y., Liang, P.S., Vaughan, J.W. (eds.)
Advances in Neural Information Processing Systems.
Curran Associates, Inc.
\end{botherref}
\endbibitem

\end{thebibliography}

\end{document}